\documentclass{modelica}
\usepackage{algorithm}
\usepackage{algpseudocode}
\algrenewcommand{\algorithmicrequire}{\textbf{Input:}}
\algrenewcommand{\algorithmicensure}{\textbf{Output:}}

\usepackage{amsmath}
\usepackage{amssymb}
\usepackage{multirow}

\hypersetup{%
  pdftitle  = {Latex Template for the International Modelica Conference},
  pdfauthor = {Author Name1, Author Name2},
  pdfsubject = {14th International Modelica Conference 2021},
  pdfkeywords = {Modelica, conference, LaTeX, template},
  colorlinks,
  linkcolor=black,
  urlcolor=black,
  citecolor=black,
  pdfpagelayout = SinglePage,
  pdfcreator = pdflatex,
  pdfproducer = pdflatex}

\begin{document}
\thispagestyle{empty}

\title{Hybrid data driven/thermal simulation model for comfort assessment}
\author[1]{Romain Barbedienne}
\author[1]{Sara Yasmine Ouerk}
\author[1]{Mouadh Yagoubi}
\author[2]{Hassan Bouia}
\author[2]{Aurélie Kaemmerlen}
\author[2]{Benoit Charrier}
\affil[1]{IRT-SystemX, France, {\small\texttt{\{romain.barbedienne,sara-yasmine.ouerk,mouadh.yagoubi\}@irt-systemx.fr}}}
\affil[2]{EDF Lab Les Renardières, EDF, France, {\small\texttt{\{hassan.bouia,aurelie.kaemmerlen,benoit.charrier\}@edf.fr}}}


\maketitle\thispagestyle{empty} 
\abstract{%
Machine learning models improve the speed and quality of physical models. However, they require a large amount of data, which is often difficult and costly to acquire. Predicting thermal comfort, for example, requires a controlled environment, with participants presenting various characteristics (age, gender, ...). This paper proposes a method for hybridizing real data with simulated data for thermal comfort prediction. The simulations are performed using Modelica Language. A benchmarking study is realized to compare different machine learning methods. Obtained results look very promising with an F1 score of 0.999 obtained using the random forest model.
}

\noindent\emph{Keywords: machine learning, hybridization, simulation, thermal comfort}

\section{Introduction}

\subsection{Context and problematic}
Nowadays, numerical simulation represents an essential tool in designing and managing real-world systems, thanks to its lower cost compared to direct experimental testing on the system to be designed. Many industrial applications have benefited from the contributions of numerical simulation to improve the performance of systems. Thermal comfort is considered as a important topic the field of numerical simulation and several studies have been conducted but the results are often far from reality \cite{feng_data-driven_2022}. One of the main difficulties is the lack of reliable data. Indeed, the acquisition of data on thermal comfort is very expensive. It requires to place the subjects in an environment where the temperature, the hygrometry rate and the thermal radiation are controlled. It also requires testing over a long period of time to avoid transient phenomena, and on a wide variety of subjects (age and gender). In this paper, we address the following problem: how can we increase the quantity of data to improve thermal comfort prediction?

\subsection{State of the art}

Time series data augmentation is a technique that aims to increase the size of the dataset using synthetic data generation or data transformation methods. This technique is used to improve the performance of time series prediction models by increasing the diversity of the training data and reducing the risk of overfitting. In the area of time series, the increase in data is particularly important because data are often scarce and expensive to collect.

\subsubsection{Data generation approaches}
One of the most well-known approaches in the field of data augmentation is synthetic data generation. Synthetic data generation approaches aim to increase the size of the dataset by generating synthetic data that resembles the real data. Some of the most common approaches include Markov processes, Gaussian mixture models and generative adversarial neural networks (GANs).

Since their inception, GANs have gained a lot of traction in the deep learning research community. Their ability to generate and manipulate data in multiple domains has contributed to their success.

A GAN is a generative model composed of a generator and a discriminator, typically two neural network (NN) models. GANs have demonstrated their ability to produce high-quality images and videos, transfer styles, and complete images. They have also been successfully used for audio generation, sequence prediction and imputation.
Jinsung Yoon et al. \cite{yoon_time-series_2019} proposed Time-series GAN (TGAN), a novel version of GAN for generating realistic time-series data. They introduced the concept of supervised loss; the model is encouraged to capture time conditional distribution within the data by using the original data as a supervision. They obtained significant improvements over state-of-the-art benchmarks in generating realistic time-series of multiple datasets.

One of the advantages of these techniques is their power to greatly increase the size of the dataset and help to model extreme situations that may not be observed in real data.

And, one of the main limitations of TGANs is the restriction of the specified sequence length that the architecture can handle. In addition, generated data may not accurately reflect real data and may require significant computational resources.

\subsubsection{Data transformation approaches}

Data transformation approaches aim to increase the size of the data set by applying transformations to the existing time series. Some common approaches include normalization, Fourier transform, time warping and interpolation.

Time warping technique consists in applying random guided transformations to existing time series to generate new training series. The time warping transformations are applied using a cost function that measures the similarity between two series. The experiments conducted in the paper \cite{iwana_time_2021} show that the proposed data augmentation technique significantly improves the performance of neural networks for various time series related tasks, such as energy consumption prediction and human activity recognition.

The Fourier Transform method \cite{yang_sfcc_2023} involves dividing the training data into multiple sets and then applying the Fourier Transform to each set. The Fourier coefficients of each set are then combined in a stratified manner to generate new training series. The newly generated series are used to train a time series classification model.

The paper also describes a method for selecting the data sets to be used for data augmentation. This method involves using a clustering algorithm to group the training data into similar sets and then selecting the data sets that are most different from each other.

Experiments conducted in the paper show that the proposed data augmentation method significantly improves the classification performance for various time series datasets.

Interpolation \cite{oh_time-series_2020} is a method of estimating unknown values in the time series using the known values based on a specific interpolation function like cubic splines. This method may greatly improve the score on the generated data especially when the interpolation function is well suited to the problem.

Other approaches using data transformation include time slicing window which consists of cutting a portion of each data sample, to generate a different new sample. Adding noise to time series, flipping by inverting a time series, scaling by changing the magnitude of a certain step in the time series, rotation and permutation. Some of these techniques can only be used for specific datasets. Indeed, it does not make sense to apply flipping for a time series describing a temperature variable for example.

One of the advantages of these techniques is their simplicity to implement and the fact that they allow to control the generated time series.

On the other hand, if the data have complex patterns the generated data may not accurately reflect the real data.

\subsubsection{Simulation approaches}

Another way for data augmentation is to use a simulation model to generate synthetic data.

For example, in the autonomous driving \cite{cao_data_2022} field, simulators such as DeepGTA-V and CARLA (Car Learning to Act) can be used to generate large amounts of synthetic data that can complement the existing real-world dataset in training autonomous car perception. These models allow to generate several scenario configurations (bad weather conditions, road accidents, obstacles...etc.), which gives different driving environments.

One potential downside of data augmentation using simulation is that the simulated data may not perfectly represent the real-world data, which can affect the performance of machine learning models trained on the data. 

On the other hand, simulation also provides more diverse information. For example, for a time series describing the operative temperature of a housing, it is possible to simulate the operative temperature in several seasons.

Simulation enrichment allow to reduce the gap between the training dataset and the dataset used for inference and evaluation. the comfort models are typically learned from real data but evaluated of these models are performs from simulation results. It is important to note that simulation results may deviate from the actual models, thus impacting the accuracy of the learning process based on simulated data. To address this issue, it is crucial to conduct the inference of the learned model using simulated variables.

However, in order to prevent any biases in machine learning (ML) models, simulations must accurately reflect the real-world context. Creating a simulation model that closely resembles the actual environment poses a significant challenge.

\section{Methodology}

\begin{figure}[h]
\centering
\includegraphics[width=0.5 \textwidth]{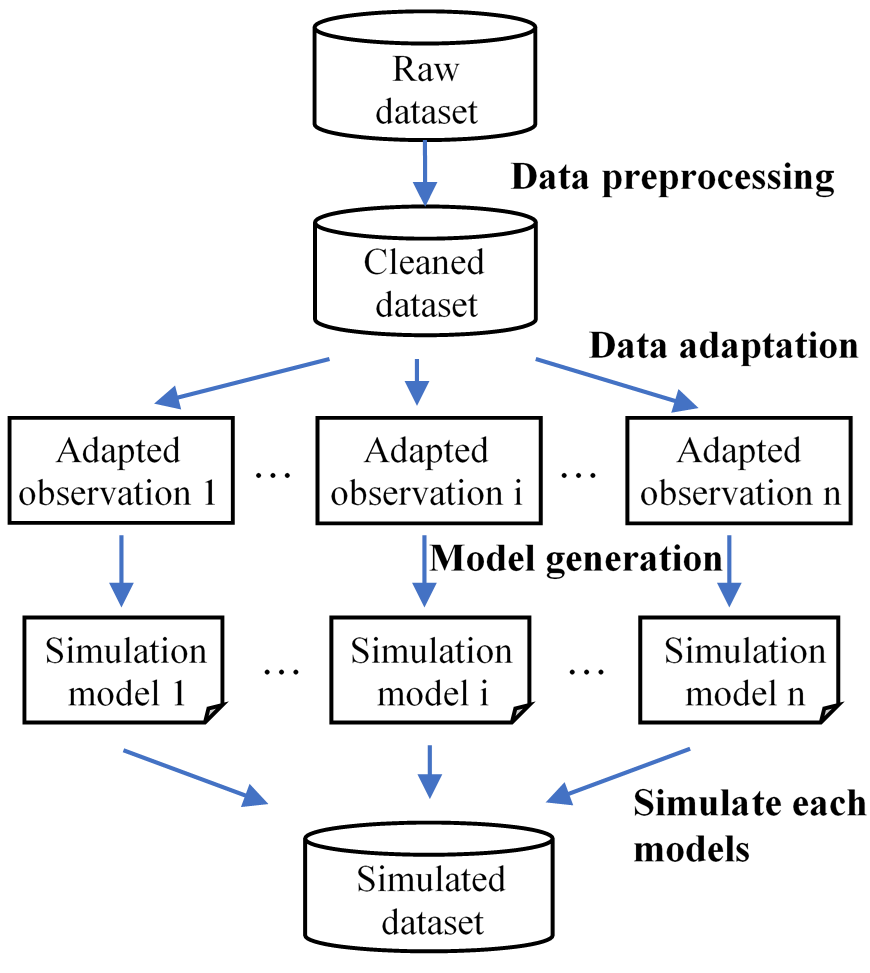}
\caption{Description of data generation process}
\label{fig:figure1}
\end{figure}

The aim of this approach is to complete each observation with environment variables generated by simulation. For each observation a simulation model will act as a digital twin. The methodology is divided into four main stages \autoref{fig:figure1}.

The first step is the data preprocessing. This is a classical step in preparing data for machine learning. The main objectives of this step are, to verify the data format, to ensure the data consistency, complete missing data and remove outliers.

The second step is the data adaptation. The prepossessed data may differ from the parameters of the model to be generated.  for example, if the data consists of questionnaires sent to a sample of people. The person answering the questionnaire may not know the parameter of a simulation model. But this parameter can be deduced from another question. For example, in the case of thermal simulation of buildings, it is easier for respondents to enter the year of construction of their house than the thickness of its insulation. It is then possible to approximate the thickness of the insulation with the norms for the year of construction. The application section \ref{sec:description} will contains more details on this step. As well as surveys.

The third step is the model generation. Model generation is performed for each observation. It requires the creation of rules to generate the simulation model, or the creation of several simulation templates whose parameters are filled in according to the adapted observations.

The last step is the simulation of each model, and the post-processing of the results. The objective of this step is to prepare simulation results for learning.

\section{Application}

\subsection{Description}\label{sec:description}

\subsubsection{Survey Analysis}
The aim of this study is to predict household thermal comfort. A survey was sent to 4000 French households. The sample was selected to be as representative as possible of the French population.

\begin{figure}[h]
\centering
\includegraphics[width=0.5 \textwidth]{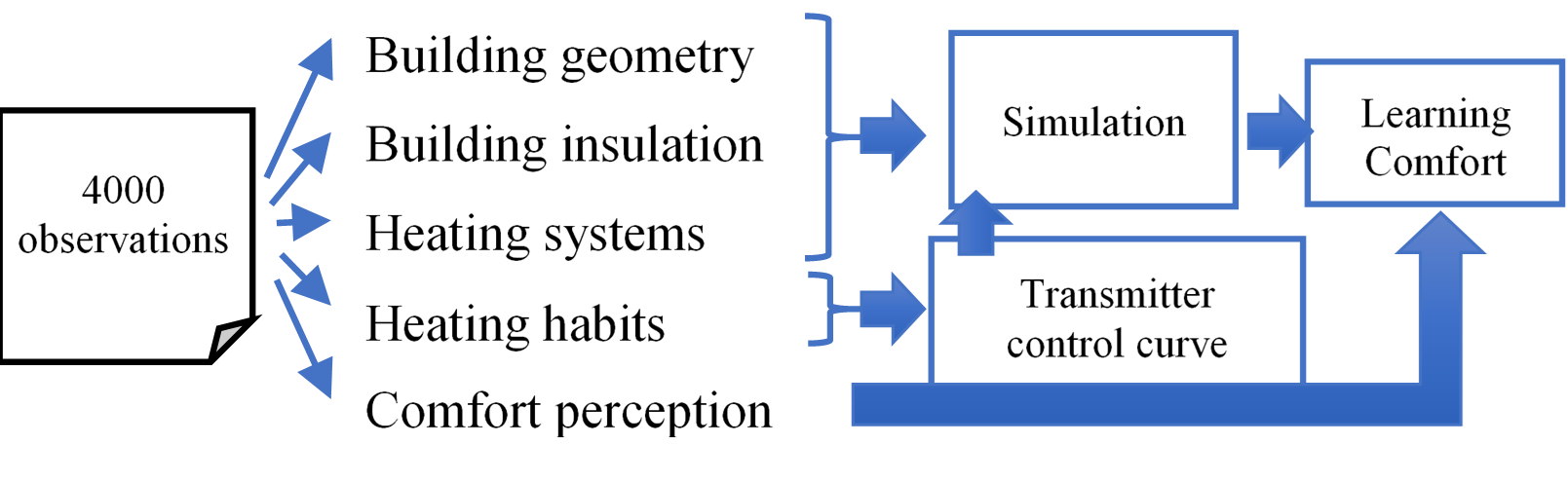}
\caption{Synopsis of approach}
\label{fig:figure2}
\end{figure}

The survey is composed of 240 questions divided into 5 categories; building geometry, building insulation, heating systems, heating habits and comfort perception. As descried in \autoref{fig:figure2}, building geometry, building insulation and heating systems questions are adapted in order to generate the thermal simulation of the housing.

 Heating habits answers are used in order to reproduce household heating curve, including, Transmitter control curve and the opening and closing curves of the shutters and windows. Finally, questions about perceived comfort are adapted to feed the learning model.
 
\subsubsection{Survey Validation And Preprocessing}

The reliability of questionnaire responses was validated by an external organization (IPSOS). Prior to the survey, 200 homes were instrumented with power and temperature sensors for each emitter. The technicians who visited the homes filled in the necessary information.  When the survey was completed, the results from these 200 homes were compared with the instrumented data to validate the approach. These results, and the comparison with simulation results, will be the subject of a future paper. The present paper deals with the methodology Process.

The first step was to pre-process the data. Dwellings containing outliers were removed. For each variable the Interquartile Rule were applied in order to identify outliers. For example, dwellings with surface too large or too small are removed. After this step, 3 529 dwellings contained statistically acceptable variables.

\subsection{Completion of data with simulation}

\subsubsection{Model hypothesis}

The simulation model used for the model generation is created thanks to buildSysPro \cite{plessis_buildsyspro_2014}. This opensource modelica library contains parametric models for different building parts, including wall, windows, roof and floor. The parametrization is simplified by grouping all the characteristic parameters in records. A record contains the parameters of the different materials used for the structure, insulation and interior cladding. It also includes geometrical parameters, such as the thickness of the different materials. The records are established according to the years of construction of the buildings and the different standards.

For every dwelling, each room is modeled as a thermal node. The temperature is considered uniform at each point of the room volume. The conductive exchanges between the walls of the rooms and the external walls are modeled using the heat transfer in one spatial dimension. Transient phenomena are considered. Walls are discretized every time materials are changed or every five cm (this is done automatically by the buildSysPro library). It allows to linearize the heat equations for each wall.

The air flow exchanges between individual rooms are neglected. We have assumed that all the interior doors are closed. Convection exchange between the outside air and the wall is calculated using the newton law. Heat transfer coefficient is given by the record according to the exterior materials.

The solar radiation is calculated using the model from Hay Davies Klucher Reindl (HDKR) \cite{padovan_measurement_2010}. The environment variables (external temperature, humidity, wind speed and direction, variables needed to calculate the incident radiative flux, etc.) are loaded from a file.

\subsubsection{Description of model templates}

\begin{figure}[h]
\centering
\includegraphics[width=0.5 \textwidth]{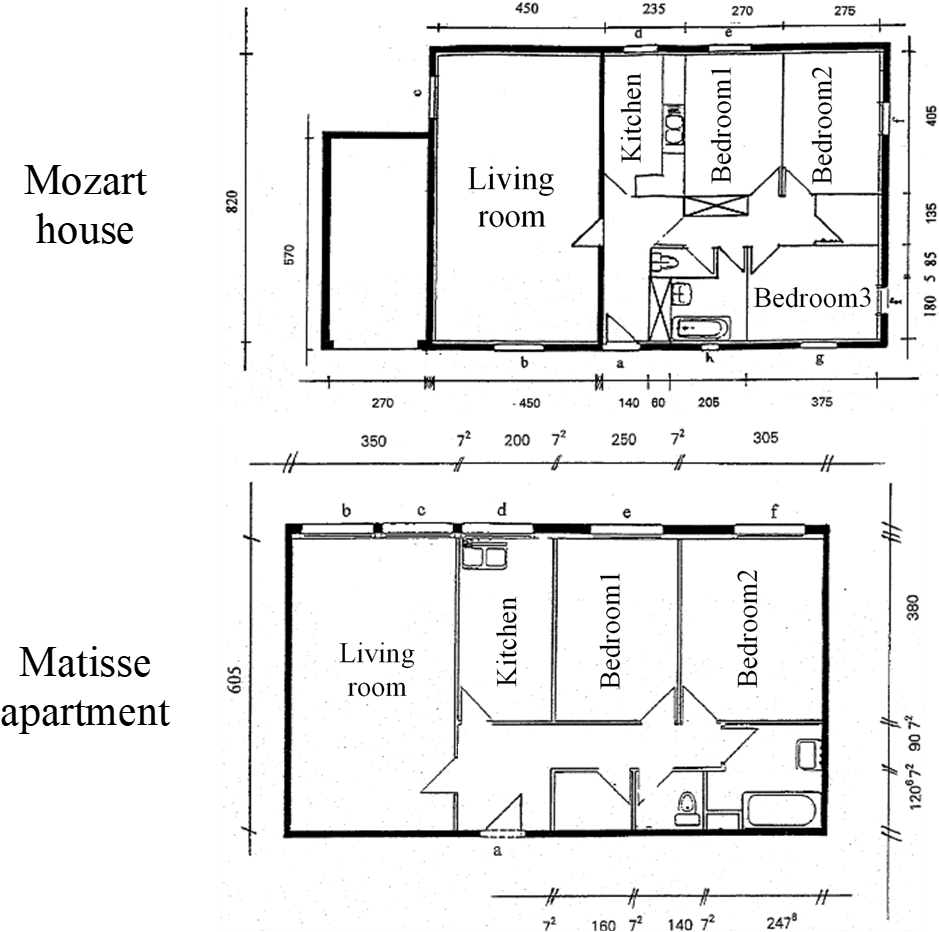}
\caption{Plan of different model templates}
\label{fig:figure3}
\end{figure}

Two generic building models are created with buildSysPro library; a 2-3 bedroom house (Mozart house) and a 1-2 bedroom apartment (Matisse apartment). 35\% of the households interviewed in the survey correspond to these two types of housing which corresponds to about 1400 dwellings. The plans of these dwellings are described in \autoref{fig:figure3}.  For each template, the Bedroom 2 can be empty.

\begin{figure}[h]
\centering
\includegraphics[width=0.5 \textwidth]{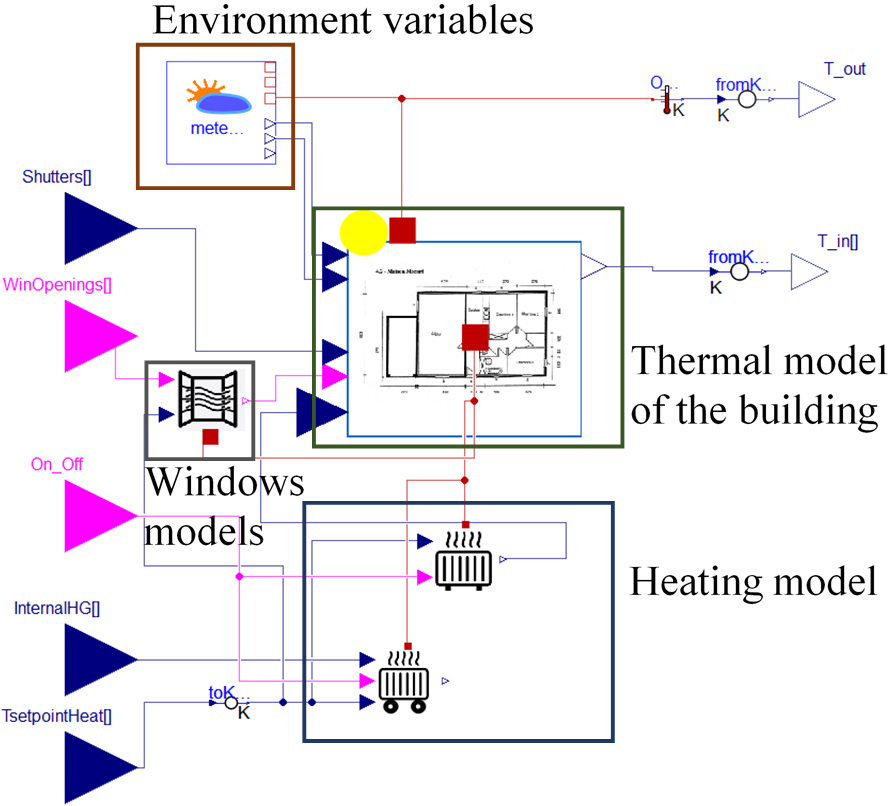}
\caption{Example of simulation template for Mozart House}
\label{fig:figure4}
\end{figure}

A simulation model template has been created by dwelling category. An example of simulation template for Mozart house is described in \autoref{fig:figure4}. In this figure, each variable is a vector. Each coordinate of the vector corresponds to a room. The dimension of the vector corresponds to the number of rooms of the dwelling. This template is composed of 4 parts; one part is the model that generates environment variables previously described. The second part is the thermal model of the building. Hypothesis of this model have also been previously described.

The window model controls the opening and closing of windows room by room. The opening of the windows is controlled by an external file (generated from the survey answers). At each time step, the model allows to open a window if its window state variable is set to true and the set temperature is reached. This model also closes the window if the difference between the set temperature and the air temperature is below a certain threshold. This threshold has been set at 3°C by default.

The heating models are composed of two models. One model controls the heat flow injected room by room by the fixed heaters. The second one controls the heat flow injected room by room by the mobile heaters. The model controlling the heat flow for mobile heaters consists in two heat flow injections; one by convection and a second one by radiation. Total heat flow injection is defined by an input csv file (generated from the survey answers). The ratio between convective and radiative heat flow is considered constant and is established according to the type of heating system entered in the questionnaire.

\begin{figure}[h]
\centering
\includegraphics[width=0.5 \textwidth]{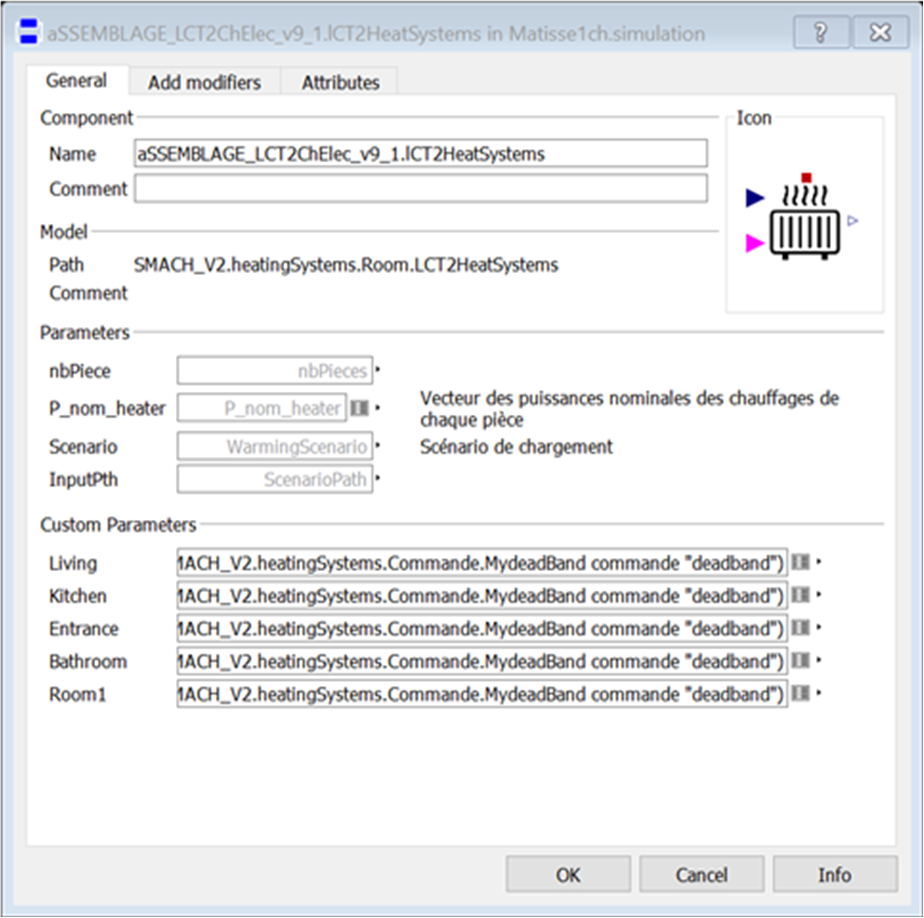}
\caption{Parametrization of fixed heaters room by room}
\label{fig:figure5}
\end{figure}

The parametrization of fixed heaters for Matisse apartment with one bedroom is described in \autoref{fig:figure5}. For each room, it is possible to set the heater type and heater controller. Variable P\_nom\_heater is a vector that contains at each coordinate the sum of the nominal power of all the heaters in a room. Scenario and InputPath parameters indicate the path for the file describing the wood reloading hours. This scenario is specific for the inhabitants using a fireplace for heating. It depends mainly on the activity of the inhabitants composing the household.

Three control models have been implemented. A Proportional integral differential (PID) model, a dead band model and a model of absence of control, when the inhabitants declared not to have heating in the room.
Six heating models have been implemented: electric heating type convector, radiant panel, soft heat, accumulation, water heating or wood heating.

Models have been implemented using method Th-BCE \cite{ministere_de_la_transition_ecologique_gestion_2023} except for wood heating model that have been instantiated from the buildSysPro library. Th-BCE method is a French regulation. The two hypothesis concerning heaters are that: the thermal inertia of heaters is neglected and the ratio of heat transfer radiated flow and convective flow is considered constant.

\subsubsection{Description of data used for simulation}

The parameters used to fill the simulation model are described in \autoref{tab:extab}. This table also describes the questions in the survey, and the method used to calculate simulation model parameters from the questions. The temporal variables were filled from the survey by choosing three typical days per week. For each day, the variables are filled in hour by hour for each room. These typical days are assigned to each day of the week.

The temperature measured in the questionnaire is based on a typical week. It does not give details of the temperature hour by hour over a year. Also, room temperatures are likely to vary according to the insulation of the dwelling and the power of the radiators. The simulation shows this variation. In the simulation, the temperature measured over a typical week is approximated as the set temperature and repeated every week. 

Two algorithms were used; a first one allows to compute the orientation variables and a second one to control the opening and closing of the shutters.

Concerning orientation variable, the chosen algorithm is a decision tree. The decision node was manually implemented. The survey contains for each room whether the windows are predominantly south-facing. Thus, orientation is calculated thanks to different plans (Matisse and Mozart) in order to maximize the surface of the windows facing to the south according to the survey. \autoref{alg:orientation} describes the decision tree for Mozart house. By convention, the Orientation variable is null when north is oriented at the top of the plan in \autoref{fig:figure3}. A similar algorithm was built for the Matisse template.

In this algorithm, ${IsSouth}_{room}$ is a binary list where $room$ belongs to each available room of Mozart house, $Orientation$ designates the variable in degree of house Orientation.

The instruction for opening and closing the shutters is calculated from the presence variable room by room. The time at which the shutters open is determined by the time at which a room becomes empty.

\begin{table}[H]
  \caption{links between questions and variables used to complete the simulation models}\label{tab:extab}
  \centering
  \begin{tabular}{|p{2.6cm}|p{2.75cm}|p{2.0cm}|} \hline
      \emph{Question} & \emph{Variable in simulation}  & \emph{Calculation method} \\
      \hline
      Number of rooms   & Number of rooms  & assignment \\
      \hline
      The total floor area of the dwelling   & Floor total area   & assignment  \\
      \hline
      Room with south-facing windows    & orientation   & deductive algorithm \\
      \hline
      Heating power for each room   & Nominal power of heater for each room   & assignment  \\
      \hline
      Year of construction of the dwelling      & House record    & assignment   \\
      \hline
      Temperature measured hour by hour over a week       & setpoint temperature   & assignment  \\
      \hline
      Ignition time of auxiliary heaters and power of auxiliary heaters & auxiliary heating power & assignment  \\
      \hline
      Hour of opening of the windows and duration of the   opening   & instruction for opening and closing the windows  & assignment   \\
      \hline
      Presence in the rooms  & instruction for opening and closing the shutters & deductive algorithm \\
      \hline
      Date of switching on the heating  & instructions for switching the heater on and off & assignment \\
      \hline
  \end{tabular}
\end{table}

The closing time of the shutters has been calculated in relation to the sunset time. The sunset time depends on the day and the location of the dwelling. The python library suntime \cite{stopa_suntime_2019} has been used for the calculation of the sunset time according to the department in which the dwelling is located and the simulated day. The closing time corresponds to the times of presence in the dwelling closest to the sunset.

\begin{algorithm}
\caption{Calculation of orientation for Mozart House}\label{alg:orientation}
\textbf{Start}
\begin{algorithmic}
\If{${IsSouth}_{living}$}
  \If{${IsSouth}_{bedroom3}$}
    \State $Orientation \gets 0^{\circ}$
  \Else
    \State $Orientation \gets 90^{\circ}$
  \EndIf
\Else
  \If{${IsSouth}_{bedroom2} \And {IsSouth}_{bedroom3}$}
    \State $Orientation \gets 270^{\circ}$
  \Else
    \State $Orientation \gets 180^{\circ}$
  \EndIf
\EndIf
\end{algorithmic}
\textbf{End}
\end{algorithm}
\subsubsection{Description of input files}

Thermal regulation 2012 (RT 2012) \cite{ministere_de_la_transition_ecologique_gestion_2023} is a French regulation. it separates France into 8 thermal zones. RT2012 provides for each thermal area, an average environment file. This file provides each 30 minutes a value for weather variables, including, wind speed and orientation, different temperatures variables, relative humidity, atmospheric pressure and solar irradiance direct and indirect. These files were used in order to estimate environment variables. The thermal area of a dwelling is determined according to its department.

\subsubsection{Generation of modelica models}

\begin{figure}[h]
\centering
\includegraphics[width=0.5 \textwidth]{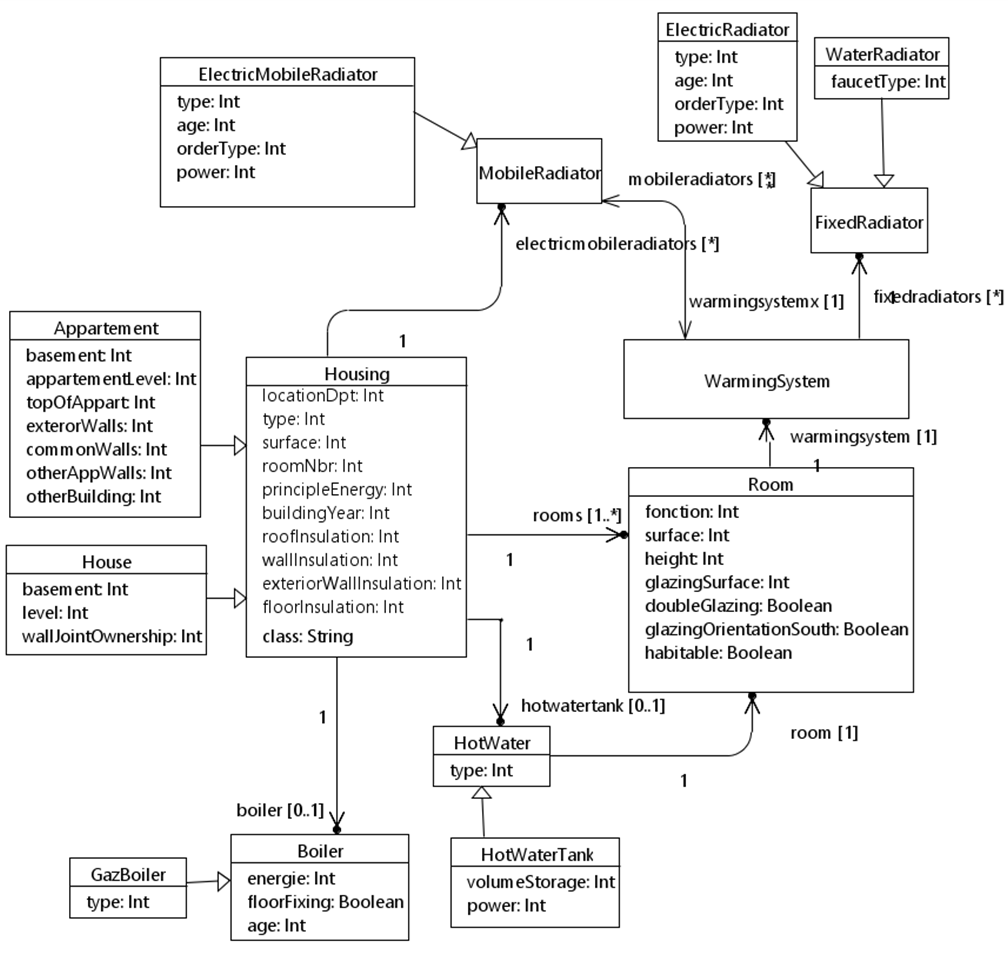}
\caption{Class Diagram of python models}
\label{fig:figure6}
\end{figure}

Each observation is instantiated using the class diagram of the algorithms as described in \autoref{fig:figure6}. Modelica files were generated from instantiation of this class diagram using model transformation process. The implementation was carried out using the Modelica Language.

The simulations were launched using Dymola-Python application programming interface (API). The solver used is Differential/Algebraic System Solver (DASSL) with a time step of 1800s. The simulations were launched from October 1st to April 30th, results files are stored as CSV files. The simulations were run on a laptop with a quad core processor with 16GB of RAM. It took 72 hours to run 1,400 simulations. Simulations were parallelized on 4 cores.

In order to facilitate the training of recurrent machine learning models, it is frequent to use an invariant time step for all the time series. Therefore, the results were post-processed to have a constant time step of 1800s, because DASSL is a variable time steps solver.

\subsubsection{Simulation results}

\begin{figure}[H]
\centering
\includegraphics[width=0.5 \textwidth]{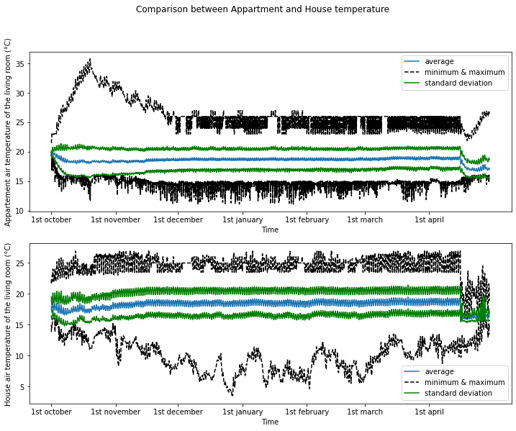}
\caption{Comparison of simulation results between House and Apartements}
\label{fig:figure7}
\end{figure}

The simulation results are given in \autoref{fig:figure7}. The average air temperature in the living room is $19^{\circ}C$ for an apartment or a house. The main difference concerns the maximum and minimum values. Indeed, the maximum and minimum temperatures are higher for apartments than for houses. The difference concerns the minimum temperature, which is explained by the proximity of an apartment to other apartments. Thus, an unheated apartment will be heated by convection by the other surrounding apartments, avoiding too low temperatures. This is not the case for houses. As regards the increase of the temperature between October 1\textsuperscript{st} and November 1\textsuperscript{st}, it concerns only one apartment, this apartment combines an early ignition of heating managed by a collective boiler with a high outside temperature due to the localization of the apartment (in the department of Corsica). Note that weather variables are approximate, and local weather conditions may be different, which explains this difference. Finally, the generalized fall of temperature after the 15\textsuperscript{th} April is due to the stop of the heating at this date.

\begin{figure}[H]
\centering
\includegraphics[width=0.5 \textwidth]{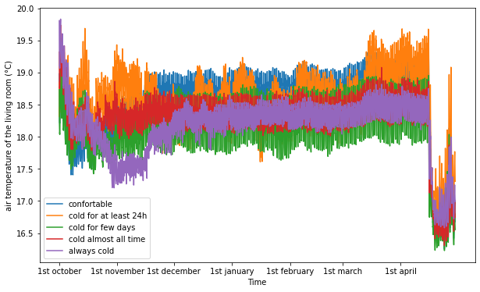}
\caption{Air temperature of living room according to comfort clusters}
\label{fig:figure8}
\end{figure}

The inhabitants surveyed were able to fill in the thermal discomfort time in the questionnaire. 5 choices were available: being comfortable (84.6\% of total observations), being cold for at least 24 hours (6.9\%), being cold for a few days (5.2\%), being cold almost all the time (1.9\%) or all the time (1.4\%). \autoref{fig:figure8} illustrates the average living room temperature of households according to their response to the comfort question. 

This figure highlights that air temperature is not a good indicator. It does not allow to differentiate between households that are cold all the time and those that are cold some days. 

The operating temperature is a good but not sufficient indicator of comfort. In fact, it considers the air temperature and the radiated temperature. However, this temperature does not represent the temperature felt by an inhabitant, because the inhabitant does not always occupy the cold room. For example, even if the operating temperature of the bathroom is $10^{\circ}C$, if there is never anyone in the bathroom it is useless to take this variable into account. Thus, we have introduced the operating temperature of presence. This variable is the operating temperature averaged by the presence of inhabitant per room.
This variable is calculated at each time step when there is at least one inhabitant in the house ($\sum_{i}^{n_{room}}{Pres}_{{room}_i}\ >\ 0$), as follow:

\begin{align}
T_{op\ pres}=\ \frac{\sum_{i}^{n_{room}}{T_{op\ {room}_i}.\ {Pres}_{{room}_i}}}{\sum_{i}^{n_{room}}{Pres}_{{room}_i}} \label{eq:a1}
\end{align}

\begin{figure}[H]
\centering
\includegraphics[width=0.5 \textwidth]{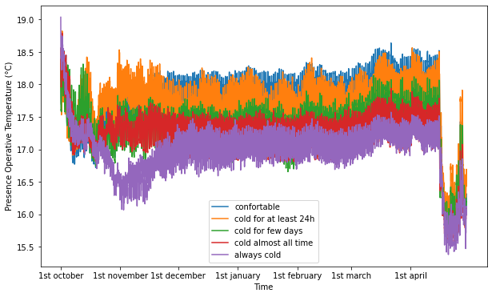}
\caption{Air temperature of living room according to comfort clusters}
\label{fig:figure9}
\end{figure}

Where; $n_{room}$ is the number of rooms for one dwelling, $T_{op\ {room}_i}$ is the operating temperature of the ${room}_i$ with $i\ \in\ 1;nroom$, ${Pres}_{{room}_i}$ is a Boolean variable for each room of one dwelling with $i\ \in\ 1;nroom$. This variable is equal to 1 if there is a presence in the room and null if there is nobody in the room and $T_{op\ pres}$ is a list composed of the operating temperatures.
As illustrated in \autoref{fig:figure9}, the presence operating temperature allows to recover the comfort trend established in the questionnaire. Therefore, we focused on this variable for learning. Note, however, that inhabitants that are cold for a few days and those that are cold almost all the time are difficult to differentiate.

\subsubsection{Preparation of data for learning}

Simulation model computes for each time step the thermal environment of the inhabitants for a dwelling. However, in order to train the ML model of thermal comfort, it is required to calculate the thermal comfort at each time step.

The first algorithm implemented is a simple threshold on the operating temperature of presence. Indeed, for each dwelling a threshold was calculated to respect the discomfort period. If operating temperature is below this threshold, the inhabitants are considered as uncomfortable, otherwise, they are considered as comfortable. The problem with this model is that there were some comfort/discomfort switches between two successive time step(s). The second issue of this approximation is that it does not consider the inertia of thermal comfort.

The improved algorithm computes for each dwelling two thresholds. When the first threshold is reached, the inhabitant is in a discomfort state. It is then required to wait for the presence operative temperature to rise to the second threshold before the inhabitant will be again considered to be comfortable. The calculation of the thresholds is performed by minimizing the number of comfort/discomfort switches under the constraint of respecting the discomfort time indicated in the survey.

For one dwelling, the problem is stated as the following optimization problem:

\begin{align}
argmin(\varepsilon_{max},n_{switch}) \label{eq:opti}
\end{align}

The constraints are:

\begin{align}
\left\{\begin{matrix}\varepsilon_{max}\ > \varepsilon_{min}\\max\left(\mathrm{\Delta}t_1,\ldots,\ \mathrm{\Delta}t_n\right)\geq t_{discomfort_survey}\\\end{matrix}\right. \label{eq:const}
\end{align}

Where; $(\varepsilon_{max},\varepsilon_{min})$ is the couple of thresholds to calculate, with $\varepsilon_{max}\in\mathbb{R}^+$ and $\varepsilon_{min}\ \in\ \mathbb{R}^+$, $n_{switch}$ is the number of comfort/discomfort switches, with $n_{switch}\ \in\mathbb{N}$, $({\Delta t}_1,\ ...,{\Delta t}_{n_{switch}}\ )$ is the list of discomfort times with ${\Delta t}_k\in\ \mathbb{R}^+$ with $k\ \in\ 1,nswitch$ and $t_{discomfort_survey}\in\ \mathbb{R}^+$ is the discomfort time indicated in the survey.

The estimation of these two thresholds is performed using the following algorithm. This algorithm is an heuristic which is divided into 3 main steps. The first step is to calculate a first value of $\varepsilon_{max}$, named $\varepsilon_0$ (\autoref{alg:firstValOfEpsi}). The calculation of $\varepsilon_0$ is similar to the calculation of a single threshold for comfort/discomfort switching. 

In this algorithm, the following variables are used; $T_{op\ pres}[\ ]$ designates a real list composed of the operating temperature at each time step, $T_{asc\_op\ pres}[\ ]$ is a real list composed of the operating temperature ordered in an ascending order, $n$ is an integer , $n_{consecutive}$ is an integer representing the number of consecutive time steps in $T_{asc\_op\ pres}$ between 0 and n, $t_{discomfort\_survey}$ a real representing the discomfort time indicated in the survey, $\Delta tstep$ is a real representing the duration of one simulation time step, $\varepsilon_0$ is a real representing the minimum threshold to have $n_{consecutive}$ discomfort time steps.

The function used are : $list\gets sorting\_asc(list)$ is a function that orders a list in an ascending order, $int\ \gets\ get\_consecutive\_tstep(int : n,list)$ is a function that returns the number of consecutive time steps in a list between $1$ and $n$.

The second step is to define a set of $\varepsilon_{max,i}$ and $\varepsilon_{min,i}$ pairs that are close to the optimal solution ${{\varepsilon_{max,1},\varepsilon_{min,1}},\ ...,{\varepsilon_{max,m},\varepsilon_{min,m}}}$ (\autoref{alg:ListofEps}). The third step is to select the optimal couple ${\varepsilon_{max,k},\varepsilon_{min,k}}$  that minimizes the objective function and respects the constraints described above.

In this algorithm, the following variables are used : $T_{op\ pres}[\ ]$ a real list composed of the operating temperature at each time step; $\varepsilon_0$ is the first value of the threshold calculated by the first algorithm; $T_{threshold_op\ pres}[\ ]$ is a real list composed of the operating temperature bellow the threshold $\varepsilon_0$; $t_{discomfort_survey}$ is a real representing the discomfort time indicated in the survey; $\Delta tstep$ is a real representing the duration of one simulation time step; $n_{clu}$ designates an integer defining the number of time steps of discomfort; $id_min[\ ]$ a list of indexes for local minimums; $i$,$minID$ ,$maxID$ are integers, $\varepsilon_{min}[\ ]$ is a list of candidates of minimal threshold value; $\varepsilon_{max}[\ ]$ is a list of candidates of maximal threshold value.

The function used are : $list\gets keep\_value\_below( list , real : threshold)$ a function that keeps the values of a list below the threshold value; $list\gets local\_minium\_list(list)$ a function allowing to calculate the local minimums of a list; $int\gets len(list)$ a function that calculates the length of the list.

\begin{algorithm}[h]
\caption{First step Find a first value for $\varepsilon_{max}$}\label{alg:firstValOfEpsi}
\textbf{Initialization}
\begin{algorithmic}
 \State $T_{asc_op\ pres}\gets\ sorting_asc(T_{op\ pres})$
 \State $n\gets{{\ t_{discomfort\_survey}} \over {\Delta t_{step}}}$
 \State $n_{consecutive} \gets get\_consecutive\_tstep(n,T_{asc\_op\ pres}\ ) $
\end{algorithmic}
\textbf{Start}
\begin{algorithmic}
  \While{$n_{consecutive}\ .\ \Delta t_{step}< t_{discomfort_survey}$}
    \State $n_{consecutive}\gets get\_consecutive\_tstep(n,T_{asc_op\ pres})$
    \State $n\gets n+1$
  \EndWhile
  \State $\varepsilon_0\ \gets\ T_{asc\_op\ pres}[n]$
\end{algorithmic}
\textbf{end}
\end{algorithm}

\begin{figure}[h]
\centering
\includegraphics[width=0.5 \textwidth]{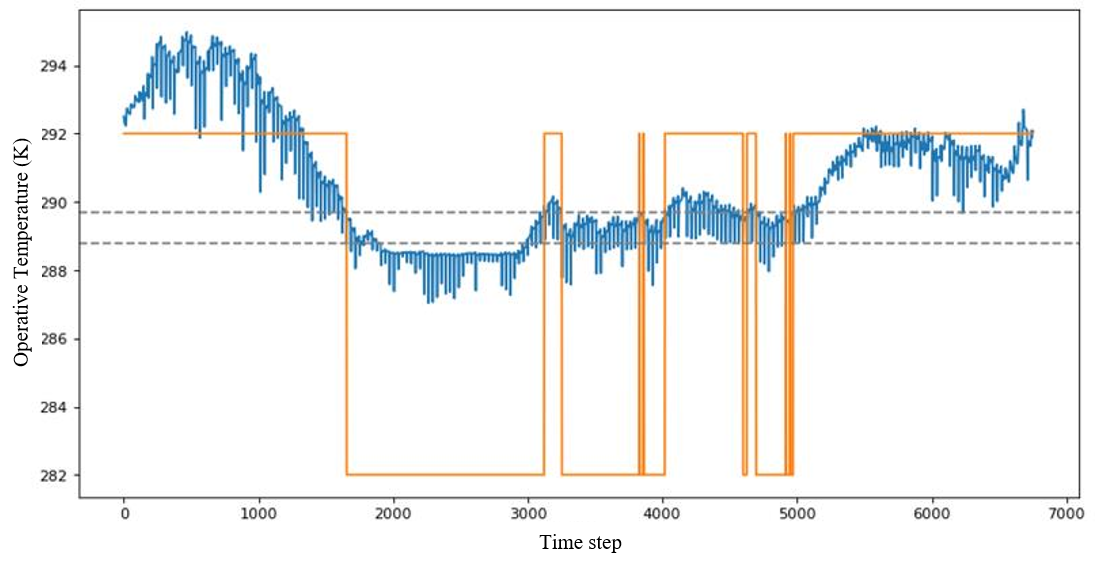}
\caption{Results of the change from comfort to discomfort in a household that is almost always cold}
\label{fig:figure10}
\end{figure}

The algorithm was implemented in python language and executed on the 1400 dwellings. \autoref{fig:figure10} shows an example for an inhabitant that reported to be cold almost all the time. In this figure, the x axis defines the time step. The y axis corresponds to the operating temperature in degrees Kelvin.

\begin{algorithm}[H]
\caption{Find candidate couples $\varepsilon_{max}$}\label{alg:ListofEps}
\textbf{Initialization}
\begin{algorithmic}
 \State $n_{clu} \gets\ {t_{discomfort_survey} /over {\Delta t_{step}}}$
 \State $i\gets 0$
\end{algorithmic}
\textbf{Start}
\begin{algorithmic}
 \State $T_{threshold\_op\ pres}\gets\ keep\_value\_below(T\_{op\ pres}\ ,\ \varepsilon_0)$
 \State $id\_min\ \gets\ local\_minium\_list(T_{threshold\_op\ pres})$
 \While{ $i < len(id\_min)$}
    \State $minID\ \gets\ id\_min[i]$
    \State $\varepsilon_{min}[i] \gets\ T_{op\ pres}[minID ]$
    \State $maxID\gets\ min(minID\ +n_{clu},len(T_{op\ pres}))$
    \State $Epsi\ \gets\ max(T_{op\ pres}[minID:maxID\ ])$
    \State $\varepsilon_{max}[i] \gets\ Epsi$
    \State $i\ \gets i\ +\ 1$
  \EndWhile
\end{algorithmic}
\textbf{End}
\end{algorithm}

\subsection{Machine learning results}

A data-driven modeling was performed in order to learn the comfort based on the both real and simulated data. The inputs considered are the simulations output at each time step (Radiation temperature, Convective temperature, presence operative temperature, heat flux emitted by radiators for each rooms, Outdoor temperature ) and sociological data from survey (average age of household, average gender of household). The output of the machine learning model is a prediction of occupant comfort and discomfort states.

The dataset was divided into 3 sets (60\% on Train, 20\% on Validation and 20\% on Test). The training set is used to train the model and calibrate its parameters, the validation set is used to prevent the model from overfitting during the training phase; by monitoring the evolution of the cost function on both sets. Finally, the test set is used the evaluate the model performances once the training is done.

A first version was built, considering the time steps independent between them. That is, the comfort at a given time step depends only on the simulated temperatures and the characteristics of the housing, and does not depend on the comfort at the time step that precedes it.

Using this configuration, several Machine Learning (ML) models were trained and tested; including Ensemble models like Random Forest and XGBoost, neural networks: Multi-Layer-Perceptron (MLP) \cite{singh_modeling_2017}.

In a second step, a new version of modeling was built. It would allow to consider that each comfort value at a given time step depends on its previous values, in addition to exogenous variables (temperatures...etc.). For this configuration, a multi-horizon model \cite{wen_multi-horizon_2018} was tested, consisting of a past horizon and a prediction horizon. This model is particularly well suited for time series prediction.

For this model, two configurations were compared. A first one was built using the real values in the past horizon of the model. The second supposes not to know these real comfort values, and therefore uses only the predicted values to feed the past horizon comfort values.

In order to train these different models, the CrossEntropy loss (CE) \cite{wang_comprehensive_2020} was used as a cost function to minimize during training, it is defined for one sample as follow:

\begin{align}
CE\ =\ \sum_{i\ =\ 1}^{C}{y_i\times l o g\left(p_i\right)} \label{eq:CE}
\end{align}

Where $C$ is the total number of classes,  $y_i$ is the truth value of the label, and $p_i$ the softmax probability for the $i^{th}$ class.

This loss penalizes the probabilities far from the truth label. The logarithm gives a large score for large differences close to 1 and small score for the ones tending to 0.  The total cost is then calculated by averaging the individual costs obtained for the different samples.

And, to evaluate and compare the different models’ performances, many classification scores were used; including precision, recall and F1 score \cite{erickson_magicians_2021} for each class of comfort. Precision represents the rate of correct predictions, recall represents the rate of positive samples detected, and the F1 score is a compromise of these two scores. These scores are defined as follow:

\begin{align}
Precision\ =\ \frac{TP}{TP+FP} \label{eq:Precision} \\
Recall\ =\ \frac{TP}{TP+FN} \label{eq:Recall} \\
F1\ =2\ \times\frac{Precision\times R e c a l l}{Precision+Recall} \label{eq:F1}
\end{align}

Where TP represents the number true positives, FP the number of false positives and FN the number of false negatives.
\autoref{tab:score} and \autoref{tab:scoreMultiH} illustrate the different classification scores evaluated on the test dataset. \autoref{tab:score} shows the results obtained using the first modeling configuration with the three models (RF, XGBoost and MLP).

\begin{table}[h]
  \caption{links between questions and variables used to complete the simulation models}\label{tab:score}
  \centering
  \begin{tabular}{p{1.5cm}p{1.3cm}p{1.2cm}p{0.8cm}p{1cm}} \toprule
    \emph{Class}  & \emph{Model} & \emph{Precision} & \emph{Recall} & \emph{F1score} \\
    \midrule
      \multirow{3}{*}{Comfort}    & MLP                                                            & 0.95      & 0.98   & 0.97    \\
                            & XGBoost                                                        & \textbf{0.999}     & \textbf{0.999}  & \textbf{0.999}   \\
                            & Random Forest & \textbf{0.999}     & \textbf{0.999}  & \textbf{0.999}   \\
                            \hline
\multirow{3}{*}{Discomfort} & MLP                                                            & 0.61      & 0.37   & 0.46    \\
                            & XGBoost                                                        & 0.97      & 0.95   & 0.96    \\
                            & Random Forest & \textbf{0.999}     & \textbf{0.999}  & \textbf{0.999}   \\
                            \hline
\multirow{3}{*}{Unknown}    & MLP                                                            & \textbf{1.0}       & \textbf{1.0}    & \textbf{1.0}     \\
                            & XGBoost                                                        & \textbf{1.0}       & \textbf{1.0}    & \textbf{1.0}     \\
                            & Random Forest & \textbf{1.0}       & \textbf{1.0}    & \textbf{1.0}    \\
      \bottomrule
  \end{tabular}
\end{table}

\autoref{tab:scoreMultiH} shows the results obtained with the second modeling configuration using the multi-horizons model with its two configurations. 
The support column represents the number of test examples used for each class of comfort.

\begin{table}[h]
  \caption{Multi-horizons model evaluation}\label{tab:scoreMultiH}
  \centering
  \begin{tabular}{p{0.5cm}p{2.01cm}p{0.75cm}p{0.75cm}p{0.75cm}p{1cm}} \toprule
    \emph{\rotatebox[origin=c]{90}{\textbf{Class}}}  & \emph{Prediction   strategy} & \emph{\rotatebox[origin=c]{90}{Precision}} & \emph{\rotatebox[origin=c]{90}{Recall}} & \emph{\rotatebox[origin=c]{90}{F1score}} & \emph{\rotatebox[origin=c]{90}{Support}} \\
    \midrule
      \multirow{2}{*}{\rotatebox[origin=c]{90}{Discomfort}} & Real   values in past horizon & \textbf{0.999}     & \textbf{0.999}  & \textbf{0.999}    & \multirow{2}{*}{33550}  \\
                            & Recursive   prediction        & 0.88      & 0.25   & 0.39     &                         \\
                            \hline
\multirow{2}{*}{\rotatebox[origin=c]{90}{Comfort}}    & Real   values in past horizon & \textbf{0.999}     & \textbf{0.999}  & \textbf{0.999}    & \multirow{2}{*}{124390} \\
                            & Recursive   prediction        & 0.83      & 0.99   & 0.90     &                         \\
                            \hline
\multirow{2}{*}{\rotatebox[origin=c]{90}{Unknown}}    & Real   values in past horizon & \textbf{1.0}       & \textbf{1.0}    & \textbf{1.0}      & \multirow{2}{*}{44220}  \\
                            & Recursive   prediction        & 1.0       & 1.0    & 1.0      &                        \\
      \bottomrule
  \end{tabular}
\end{table}

From \autoref{tab:score}, with the first configuration, the random forest model performed very promisingly for all three comfort classes with the different evaluation metrics. This is likely due to the fit between how the comfort labels were defined using the thresholds and how a decision tree (unit of an RF model) works. In addition, a random forest model is composed of a set of simple decision trees, making it accurate and robust on small datasets.

On the other hand, \autoref{tab:scoreMultiH} shows that with the second configuration, the multi-horizon model can also obtain very promising results when it is possible to feed its past horizon with the actual comfort values. Unfortunately, for this use case, and with the available data, this configuration cannot be applied because the comfort values in the past horizon cannot be available for each time step.
Therefore, according to these benchmark results, the most sweated model for comfort modeling is the random forest model which is simple and demonstrated very accurate prediction results.

\section{Conclusion}

The hybridization of thermal simulation and data-based modeling addressed the problem of data scarcity and allowed for the inclusion of additional variables not captured in the survey. Various machine learning models were trained and tested, with the random forest model performing best. 

This first study considers temperature, convective, and radiative flux variables. To improve the accuracy and realism of the approach, humidity and air speed parameter shall be considered. In this context, implementing a Stolwijk model \cite{stolwijk_mathematical_1971} instead of calculating thresholds on operating temperature would significantly improve the realism of simulated data. Additionally, integrating a multi-agent model like the SMACH model \cite{albouys_smach_2019} developed by EDF would help for making more accurate predictions of comfort. Finally, a more complete simulation model, modeling air exchanges between each room, could improve the precision of results.

Lastly, although this approach is promising, it has taken a long time to develop. A comparison of the cost, quality, development time and repeatability of the different approaches would allow to assess which approach is best suited to the need. The LIPS platform \cite{leyli_abadi_lips-learning_2022} will be used to perform such a benchmark.

\section{Acknowledgements}

This work was supported by funds from the French Program ”Investissements d’Avenir”.

\printbibliography

@article{feng_data-driven_2022,
	title = {Data-driven personal thermal comfort prediction: {A} literature review},
	volume = {161},
	journal = {Renewable and Sustainable Energy Reviews},
	author = {Feng, Yanxiao and Liu, Shichao and Wang, Julian and Yang, Jing and Jao, Ying-Ling and Wang, Nan},
	year = {2022},
	note = {Publisher: Elsevier},
	pages = {112357},
}

@article{yoon_time-series_2019,
	title = {Time-series generative adversarial networks},
	volume = {32},
	journal = {Advances in neural information processing systems},
	author = {Yoon, Jinsung and Jarrett, Daniel and Van der Schaar, Mihaela},
	year = {2019},
}

@inproceedings{iwana_time_2021,
	title = {Time series data augmentation for neural networks by time warping with a discriminative teacher},
	booktitle = {2020 25th {International} {Conference} on {Pattern} {Recognition} ({ICPR})},
	publisher = {IEEE},
	author = {Iwana, Brian Kenji and Uchida, Seiichi},
	year = {2021},
	pages = {3558--3565},
}

@article{yang_sfcc_2023,
	title = {{SFCC}: {Data} {Augmentation} with {Stratified} {Fourier} {Coefficients} {Combination} for {Time} {Series} {Classification}},
	volume = {55},
	number = {2},
	journal = {Neural Processing Letters},
	author = {Yang, Wenbo and Yuan, Jidong and Wang, Xiaokang},
	year = {2023},
	note = {Publisher: Springer},
	pages = {1833--1846},
}

@article{oh_time-series_2020,
	title = {Time-series data augmentation based on interpolation},
	volume = {175},
	journal = {Procedia Computer Science},
	author = {Oh, Cheolhwan and Han, Seungmin and Jeong, Jongpil},
	year = {2020},
	note = {Publisher: Elsevier},
	pages = {64--71},
}

@article{cao_data_2022,
	title = {Data generation using simulation technology to improve perception mechanism of autonomous vehicles},
	journal = {arXiv preprint arXiv:2207.00191},
	author = {Cao, Minh and Ramezani, Ramin},
	year = {2022},
}

@article{padovan_measurement_2010,
	title = {Measurement and modeling of solar irradiance components on horizontal and tilted planes},
	volume = {84},
	number = {12},
	journal = {Solar Energy},
	author = {Padovan, Andrea and Del Col, Davide},
	year = {2010},
	note = {Publisher: Elsevier},
	pages = {2068--2084},
}

@inproceedings{plessis_buildsyspro_2014,
	title = {{BuildSysPro}: a {Modelica} library for modelling buildings and energy systems},
	booktitle = {Proceedings of the 10 th {International} {Modelica} {Conference}; {March} 10-12; 2014; {Lund}; {Sweden}},
	publisher = {Linköping University Electronic Press},
	author = {Plessis, Gilles and Kaemmerlen, Aurélie and Lindsay, Amy},
	year = {2014},
	note = {Issue: 096},
	pages = {1161--1169},
}

@misc{ministere_de_la_transition_ecologique_gestion_2023,
	title = {Gestion des versions du {Moteur} de calcul “ th-{BCE} 2020 ” et du {RSEE}},
	url = {https://rt-re-batiment.developpement-durable.gouv.fr/gestion-des-versions-du-moteur-de-calcul-th-bce-a688.html},
	author = {de la transition écologique Ministère},
	year = {2023},
	note = {Publication Title: RT-RE-bâtiment},
}

@misc{stopa_suntime_2019,
	title = {Suntime},
	url = {https://pypi.org/project/suntime/},
	urldate = {2023-06-09},
	author = {Stopa, Krzysztof},
	year = {2019},
	note = {Publication Title: PyPI},
}

@article{singh_modeling_2017,
	title = {Modeling the spatial dynamics of deforestation and fragmentation using {Multi}-{Layer} {Perceptron} neural network and landscape fragmentation tool},
	volume = {99},
	journal = {Ecological Engineering},
	author = {Singh, Sonali and Reddy, C Sudhakar and Pasha, S Vazeed and Dutta, Kalloli and Saranya, KRL and Satish, KV},
	year = {2017},
	note = {Publisher: Elsevier},
	pages = {543--551},
}

@article{wen_multi-horizon_2018,
	title = {A {Multi}-{Horizon} {Quantile} {Recurrent} {Forecaster}},
	author = {Wen, Ruofeng and Torkkola, Kari and Narayanaswamy, Balakrishnan and Madeka, Dhruv},
	year = {2018},
	note = {\_eprint: 1711.11053},
}

@article{wang_comprehensive_2020,
	title = {A comprehensive survey of loss functions in machine learning},
	journal = {Annals of Data Science},
	author = {Wang, Qi and Ma, Yue and Zhao, Kun and Tian, Yingjie},
	year = {2020},
	note = {Publisher: Springer},
	pages = {1--26},
}

@misc{erickson_magicians_2021,
	title = {Magician’s corner: 9. {Performance} metrics for machine learning models},
	publisher = {Radiological Society of North America},
	author = {Erickson, Bradley J and Kitamura, Felipe},
	year = {2021},
	note = {Issue: 3
Pages: e200126
Publication Title: Radiology: Artificial Intelligence
Volume: 3},
}

@article{leyli_abadi_lips-learning_2022,
	title = {{LIPS}-{Learning} {Industrial} {Physical} {Simulation} benchmark suite},
	volume = {35},
	journal = {Advances in Neural Information Processing Systems},
	author = {Leyli Abadi, Milad and Marot, Antoine and Picault, Jérôme and Danan, David and Yagoubi, Mouadh and Donnot, Benjamin and Attoui, Seif and Dimitrov, Pavel and Farjallah, Asma and Etienam, Clement},
	year = {2022},
	pages = {28095--28109},
}

@techreport{stolwijk_mathematical_1971,
	title = {A mathematical model of physiological temperature regulation in man},
	institution = {NASA},
	author = {Stolwijk, Jan AJ},
	year = {1971},
}

@inproceedings{albouys_smach_2019,
	title = {{SMACH}: {Multi}-agent {Simulation} of {Human} {Activity} in the {Household}},
	booktitle = {Advances in {Practical} {Applications} of {Survivable} {Agents} and {Multi}-{Agent} {Systems}: {The} {PAAMS} {Collection}: 17th {International} {Conference}, {PAAMS} 2019, Ávila, {Spain}, {June} 26–28, 2019, {Proceedings} 17},
	publisher = {Springer},
	author = {Albouys, Jérémy and Sabouret, Nicolas and Haradji, Yvon and Schumann, Mathieu and Inard, Christian},
	year = {2019},
	pages = {227--231},
}

\end{document}